\documentclass[10pt,twocolumn,letterpaper]{article}

\usepackage{cvpr}              

\usepackage{graphicx}
\usepackage{amsmath}
\usepackage{amssymb}
\usepackage{booktabs}
\usepackage{comment}
\usepackage{multirow}
\usepackage{color}

\usepackage[pagebackref,breaklinks,colorlinks]{hyperref}

\usepackage[capitalize]{cleveref}
\crefname{section}{Sec.}{Secs.}
\Crefname{section}{Section}{Sections}
\Crefname{table}{Table}{Tables}
\crefname{table}{Tab.}{Tabs.}

\def\cvprPaperID{4025} 
\def\confName{CVPR}
\def\confYear{2022}

\begin{document}

\title{Oriented RepPoints for Aerial Object Detection}

\author{Wentong Li$^1$, \ \ \  Yijie Chen$^1$,  \ \ \ Kaixuan Hu$^2$, \ \ \ Jianke Zhu$^{1,3}\thanks{Corresponding author is Jianke Zhu.}$\\
	$^1$Zhejiang University \ \ \ \ \ \ \ $^2$University of Electronic Science and Technology of China\\
	$^3$Alibaba-Zhejiang University Joint Research Institute of Frontier Technologies \\ 
	{\tt\small \{liwentong, chen\_yj, jkzhu\}@zju.edu.cn,
		201922100522@std.uestc.edu.cn}
}
\maketitle

\begin{abstract}
    In contrast to the generic object, aerial targets are often non-axis aligned with arbitrary orientations having the cluttered surroundings. Unlike the mainstreamed approaches regressing the bounding box orientations, this paper proposes an effective adaptive points learning approach to aerial object detection by taking advantage of the adaptive points representation, which is able to capture the geometric information of the arbitrary-oriented instances. To this end, three oriented conversion functions are presented to facilitate the classification and localization with accurate orientation. Moreover, we propose an effective quality assessment and sample assignment scheme for adaptive points learning toward choosing the representative oriented reppoints samples during training, which is able to capture the non-axis aligned features from adjacent objects or background noises. A spatial constraint is introduced to penalize the outlier points for roust adaptive learning. Experimental results on four challenging aerial datasets including DOTA, HRSC2016, UCAS-AOD and DIOR-R, demonstrate the efficacy of our proposed approach. The source code is availabel at: \href{https://github.com/LiWentomng/OrientedRepPoints}{https://github.com/LiWentomng/OrientedRepPoints}.
\end{abstract}

\section{Introduction}
\label{sec:intro}
Being an important computer vision task~\cite{DBLP:conf/cvpr2018/DOTA,iccv2019clustered,ding2021dota-pami}, aerial object detection has recently attracted increasing attention, which plays the significant role in the remote image understanding. Different from the generic object detection,  aerial target localization has its own difficulties, including non-axis aligned objects with arbitrary orientations ~\cite{DBLP:conf/cvpr2019/RoI-Transformer,CVPR2020-DRN,cvpr2021redet} and densely packed distribution with complex context~\cite{DBLP:conf/iccv2019/SCRDet,iccv2019clustered, CVPR2021beyond}.

\begin{figure}[t]
	\centering
	\begin{subfigure}{0.44\textwidth}
	\includegraphics[width=\textwidth]{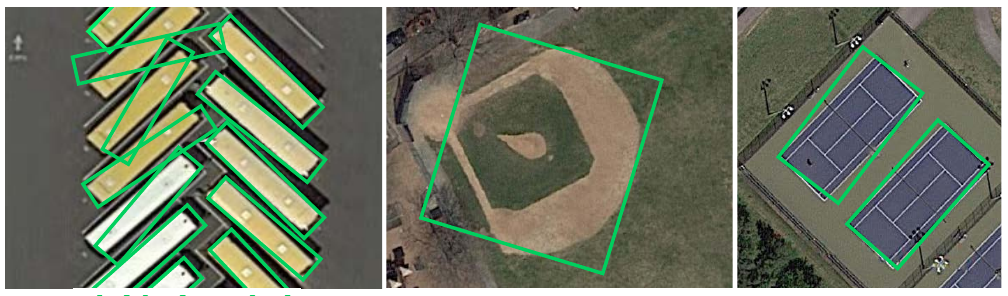}
	\caption{Orientation-regressed method-RetinaNet~\cite{DBLP:conf/iccv2017/FocalLoss}}
	\end{subfigure}
	
	\begin{subfigure}{0.44\textwidth}
	\includegraphics[width=\textwidth]{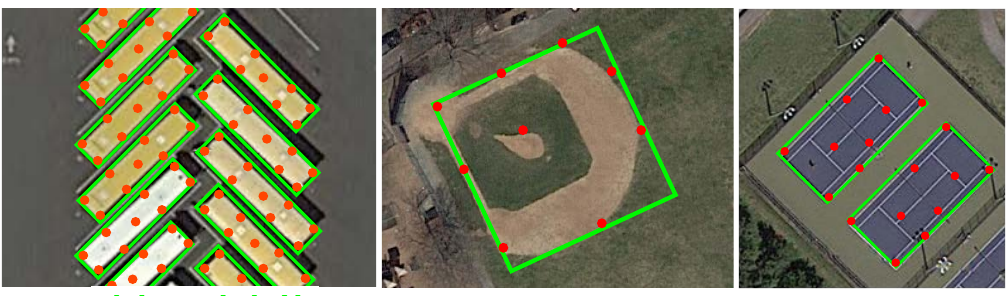}
	\caption{Proposed Oriented RepPoints framework}
	\end{subfigure}
	\caption{(a) denotes the common baseline-RetinaNet~\cite{DBLP:conf/iccv2017/FocalLoss} of orientation regression-based methods, (b) is the baseline method of our Oriented RepPoints. Comparing to the direct orientation regression, our approach can estimate more accurate orientations by learning the adaptive points that are marked as red.}\label{methods_compare}
\end{figure}

The mainstreamed approaches typically treat aerial object detection as a problem of rotated object localization~\cite{DBLP:conf/iccv2019/SCRDet,DBLP:conf/cvpr2019/RoI-Transformer,AAAI2020r3det,CVPR2020-DRN,cvpr2021redet,s2anet-2020}. Among them, the direct angle-based orientation regression methods dominate this research area, which are derived from the general vanilla detectors~\cite{DBLP:conf/nips2015/FasterRCNN, DBLP:conf/cvpr2017/FPN, DBLP:conf/iccv2017/FocalLoss, centernet2019objects} with the extra orientation parameter. Although having achieved promising performance, the direct orientation prediction still has some issues including the discontinuity of loss and regression inconsistency~\cite{AAAI2020-RSdet, CSL_ECCV2020,DCLCVPR2021,ICML2021GWD}. This is mainly due to the bounded periodic nature of the angular orientation and the orientation definition of the rotated bounding box. Despite of their attractive localization results, the orientation regression-based detectors may not accurately predict the orientations, as shown in Fig.~\ref{methods_compare}-(a).

To effectively address the above issues, we revisit the representation for aerial objects in order to avoid the sensitive orientation estimation. As a fine-grained object representation, point set shows the great potential to capture the key semantic features in conventional generic detector like RepPoints~\cite{DBLP:conf/iccv2019/RepPoint}. However, its simple conversion functions only produce the upright-horizontal bounding boxes, which cannot precisely estimate the aerial objects' orientations. Moreover, RepPoints only regresses the key points according to the semantic features while ignoring to effectively measure the quality of learned points. This may lead to the inferior performance for the non-axis aligned objects with dense-distribution and complex scene in aerial images.

In this work, we proposed an oriented object detector for aerial images, named \textit{Oriented RepPoints}, which introduces the adaptive points representation for diverse orientations, shapes and poses. 
In contrast to conventional orientation regression approach, our proposed method not only achieves the precise aerial detection with accurate orientation, but also captures the underlying geometric structure of the arbitrary-oriented aerial instances, as shown in Fig.~\ref{methods_compare}. Specifically, the initial adaptive points are generated from the center point, which are further refined to adapt for the aerial object. To obtain the oriented bounding box, three oriented conversion functions are presented according to the layouts of the learned points. Moreover, an effective adaptive points assessment and assignment (APAA) scheme is proposed for  point set learning, which measures the quality of oriented reppoints not only from the classification, localization but also from their orientation and point-wise feature correlation during training. Such scheme enables the detector to capture the non-axis aligned features from adjacent objects or background noises toward assigning the representative oriented reppoints samples. Furthermore, a spatial constraint is proposed to facilitate the vulnerable points to find their instance owner from the complex context in the aerial scene. Compared with the orientation regression-based methods, our framework obtains more precise detection performance with accurate orientation.

In summary, the main contributions of this paper are: (1) an effective aerial object detector named Oriented RepPoints, where the flexible adaptive points are introduced as the representation to achieve the oriented object detection; (2) a novel quality assessment and sample assignment scheme for adaptive points learning, which selects the points sample not only from the classification, localization but also from the orientation, point-wise feature correlation; (3) extensive experiments on four challenging datasets showing promising qualitative and quantitative results.



\section{Related Work}

\begin{figure*}[t]
	\centering
	\includegraphics[width=0.90\linewidth]{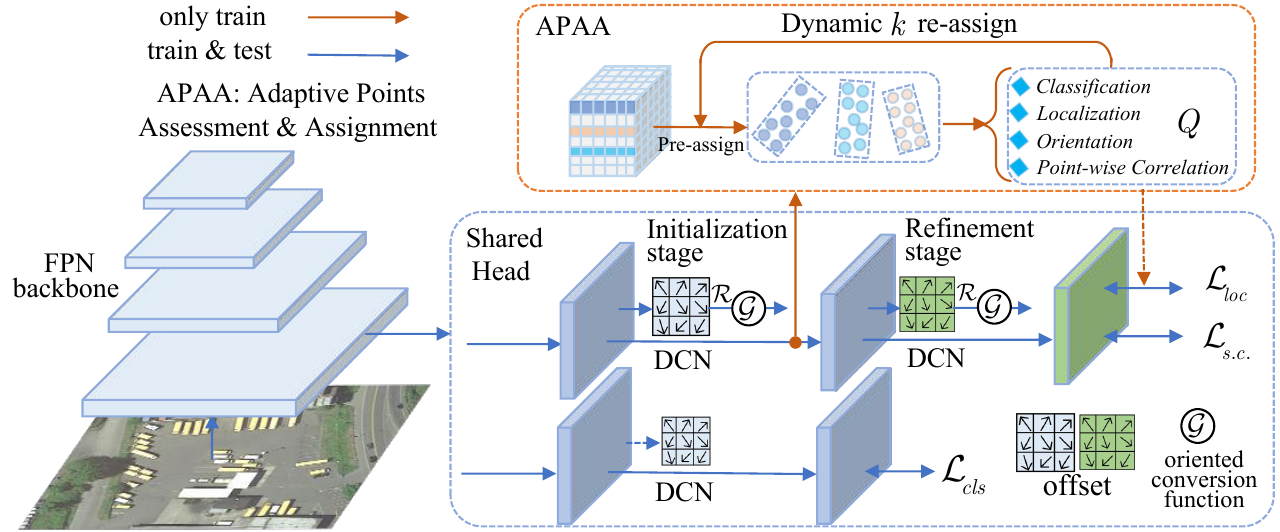}
	\caption{The framework of Oriented RepPoints. The proposed method is an anchor-free approach with the adaptive points as the representation, where a backbone with FPN network is employed for feature encoding. The structure of the shared head is same as RepPoints~\cite{DBLP:conf/iccv2019/RepPoint} for each scale of FPN, except of the proposed APAA and the oriented conversion function.	
	Based on learning points from the initialization stage, the APAA scheme is performed only during training.}
	\label{fig:overallnetwork}
\end{figure*}
Unlike most of generic object detectors with horizontal bounding boxes, the targets in aerial images are often arbitrary-oriented and dense-distributed. We discuss the related work in the following.
\subsection{Oriented Object detection}
The recent aerial object detection methods are mainly derived from the classical object detectors by introducing the orientation regression. SCRDet~\cite{DBLP:conf/iccv2019/SCRDet}, CADNet~\cite{TGRS2019-CAD}, DRN~\cite{CVPR2020-DRN}, R3Det~\cite{AAAI2020r3det}, ReDet~\cite{cvpr2021redet} and Oriented R-CNN~\cite{ICCV2021_orientedrcnn} achieve the promising performance by predicting the rotation angles of bounding boxes. Gliding Vertex~\cite{TPMI2020-Gliding} and RSDet~\cite{AAAI2020-RSdet} improve the detection results through regressing quadrilateral. To address the boundary discontinuity in the angel-based orientation estimation, Yang~\textit{et al.}~\cite{CSL_ECCV2020} transform the angular regression to angular classification~\cite{DCLCVPR2021}. Later, Yang~\textit{et al.}~\cite{ICML2021GWD} convert the parameterization of the rotated bounding box into 2-D Gaussian distributions, which gains more robust results for the oriented object detection. These methods are devoted to improving orientation estimation using rotation angle representation. Alternatively, we introduce a more effective representation using adaptive points in this paper. 

\subsection{Non-axis Aligned Features Learning}
Most of conventional object detection methods~\cite{DBLP:conf/nips2015/FasterRCNN, redmon2016you, DBLP:conf/iccv2019/RepPoint, fcos_iccv2019, centernet2019objects, zhang2020attribute,zhang2020feature} focus on either upright or axis-aligned objects, which may have difficulties with the non-axis aligned targets densely distributed in the complex background. To address this issue, Ding~\textit{et al.}~\cite{DBLP:conf/cvpr2019/RoI-Transformer} adopt the spatial transformations on axis-align RoIs and learn the non-axis aligned representation under the supervision of oriented bounding box. SCRDet++~\cite{yang2020scrdet++} enhances the non-axis aligned features and bring the higher object response to train the network. Han~\textit{et al.}~\cite{s2anet-2020} design a feature alignment module to alleviate the misalignment between axis-aligned convolutional
features and arbitrary oriented objects. 
DRN~\cite{CVPR2020-DRN} proposes the feature selection module to aggregate the non-axis aligned information from the different kernel sizes, shapes and orientations, and employs the dynamic filter generator for further regression.  Guo~\textit{et al.}~\cite{CVPR2021beyond} employ the convex-hull representation to learn the irregular shapes and layouts, which intend to avoid the feature aliasing via learnable feature adaption. Our point set-based method amis to capture the key features for non-axis aligned aerial objects.





\subsection{Samples Assignment for Object Detection}
A large amount of detection methods adopt a simple way to set the IoU threshold for selecting the positive samples. 
However, such scheme cannot guarantee the overall quality of training samples due to the potential noise and hard cases~\cite{cvpr2020learningfromnoisy, macvpr2021iqdet}. Some recent  samples assignment methods in general object detection, such as the ATSS~\cite{cvpr2020atss}, FreeAnchor~\cite{nips2019freeanchor},  PAA~\cite{eccv2020paa} and OTA~\cite{cvpr2021ota}, employ a learning-to-match optimization strategy~\cite{tpami2021learningmatching} to choose the high-quality samples. In aerial scene,  it is essential to select the high-quality samples for learning the oriented detector due to the diversity of the orientation  and dense distribution. Ming~\textit{et al.}~\cite{aaai2021dynamic} introduce a matching degree measure to evaluate the spatial alignment based on the oriented anchors, which use the matching sensitive loss to enhance the correlation between classification and oriented localization.  
In this work, we suggest an effective quality assessment and sample assignment scheme to select the positive points samples. 

\section{Oriented RepPoints}

\subsection{Overview}




Instead of directly regressing the orientations like the conventional methods~\cite{DBLP:conf/iccv2019/SCRDet,cvpr2021redet,DBLP:conf/cvpr2019/RoI-Transformer,s2anet-2020}, we take advantage of the adaptive point set~\cite{DBLP:conf/iccv2019/RepPoint} as a fine-grained representation, which is able to capture the geometric structure of the aerial objects with sharp variety on orientation in the cluttered environments. To this end, we introduce the differentiable conversion functions, where the representative points are driven to adaptively move toward the appropriate positions over an oriented object. In order to effectively learn the high-quality adaptive points without direct points-to-points supervision, we suggest a quality measure scheme that selects the high-quality oriented reppoints at the training stage. To facilitate the robust adaptive point learning, the spatial constraints are employed to penalize the vulnerable outliers and find their instance owner from the complicated aerial context. Fig.~\ref{fig:overallnetwork} shows the overview of our proposed Oriented Reppoints approach.

\subsection{Adaptive Points Learning with Orientation}

To facilitate the oriented detector with point set representation,  the conversion function is introduced to transform the adaptive points into the oriented bounding box. Let $\mathcal{G}$ denote the oriented conversion function as below:

\begin{equation}
OB = \mathcal{{{G}}({R})}
\end{equation}
where $OB$ represents an oriented box converted from the learned point set $\mathcal{{R}}$.
In this paper, we examine three oriented conversion functions:

$\bullet$ $\textit{MinAeraRect}$ intends to find the rotated rectangle with the minimum area from the learned point set over an oriented object.

$\bullet$ $\textit{NearestGTCorner}$ makes use of the ground-truth annotations. For each corner, we find the closest point from the learned point set as a predicted corner, where the selected corner points are used to build a quadrilateral as the oriented bounding box.

$\bullet$ $\textit{ConvexHull.}$ An oriented instance polygon can be defined as a convex hull of a set of points drived by the Jarvis March algorithm~\cite{jarvis1973convexhull, CVPR2021beyond}, which is used by many contour-based methods.

Note that the $\textit{NearestGTCorner}$ and $\textit{ConvexHull}$ are differentiable functions while $\textit{MinAeraRect}$ is not. Thus, we employ $\textit{MinAeraRect}$ in the post-processing to get the standard rotated rectangle prediction, and the other two differentiable functions are used to optimize adaptive points learning during the training. Under the supervision of the oriented ground-truth annotations, the points move towards the semantic key and geometric features adaptively for each aerial object, which are driven by the classification and localization loss simultaneously. 

The proposed framework consists of two stages. The initialization stage generates the adaptive point sets by refining from the object center point (feature map bins). The refinement stage further gains the accurate adjustment by minimizing the loss function as below:
\begin{equation}
\mathcal{L} = {\mathcal{L}_{cls}} + {\lambda _1}{\mathcal{L}}_{s1} + {\lambda _2}{\mathcal{L}}_{s2}
\label{overallloss}
\end{equation}
where ${\lambda _1}$ and ${\lambda _2}$ are balanced weighting. 
 $\mathcal{L}_{cls}$ denotes the object classification loss:
\begin{equation}
{{\mathcal{L}}_{cls}} = \frac{1}{{{N_{cls}}}}\sum\limits_i {{{\mathcal{F}}_{cls}}} (\mathcal{R}_i^{cls}(\theta ),b_j^{cls})
\end{equation}
where $\mathcal{R}_i^{cls}(\theta )$ represents the predicted class confidence based on the learned points, and $b_j^{cls}$ is the assigned ground-truth class. ${{\mathcal{F}_{cls}}}$ is the focal loss~\cite{DBLP:conf/iccv2017/FocalLoss}. ${{N_{cls}}}$ denotes the total number of point sets. $\mathcal{L}_{s1}$ and  $\mathcal{L}_{s2}$ represent the spatial localization loss at the initialization and refinement stage, respectively. For each stage, $\mathcal{L}_{s}$ can be denoted as below:
\begin{equation}
{{\mathcal{L}}_s} = \mathcal{L}_{loc} + \mathcal{L}_{s.c.}
\end{equation}
where ${{\mathcal{L}_{loc}}}$ is localization loss based on converted oriented boxes, and $\mathcal{L}_{s.c.,}$ denotes the spatial constraint loss.

Let ${{N_{loc}}}$ denote the total number of positive point set samples. $b_j^{loc}$ indicates the location of ground-truth box. Thus, $\mathcal{L}_{loc}$ is defined as follows:
\begin{equation}
{{\mathcal{L}}_{loc}} = \frac{1}{{{N_{loc}}}}\sum\limits_i {[b_j^{cls} \ge 1]{{\mathcal{F}}_{loc}}} ({OB_i^{loc}}(\theta ),b_j^{loc})
\end{equation}
where ${\mathcal{F}}_{loc}$  is the GIoU loss~\cite{giou-cvpr2019} for the oriented polygon.

Due to the diversity of different categories and the cluttered background in aerial images, a portion of learned points are susceptible to the background or adjacent objects with strong key features, which may move outside the ground-truth bounding box. To facilitate the vulnerable points to capture the geometric features on its instance owner, we introduce an effective spatial constraint to penalize the adaptive points outside the bounding box. Let $\rho_{ij}$ denote the penalty function. The spatial loss $\mathcal{L}_{s.c.}$ for each oriented object is defined as below:
\begin{equation}
\mathcal{L}_{s.c.} = \frac{1}{{{N_a}}} \frac{1}{{{N_o}}} \sum\limits_{i=1} {\sum\limits_{j= 1} {{\rho _{ij}}} } 
\end{equation}
where $N_a$ indicates the number of assigned positive point set samples for each object. $N_o$ is the number of points outside the GT box in each point set.

Let ${\mathbf{p}_{c}}$ denote the geometric center of the ground-truth bounding box. Given a sampled point ${\mathbf{p}_{o}}$ outside the bounding box, the penalty term is defined as below:
\begin{equation}
{\rho} =\left\{\begin{array}{l}
\|{\mathbf{p_o}}-{\mathbf{p}_{c}}\|,\ \ \ \ {\mathbf{p}_o} \  \mathrm{is\ outside} \ GT\\
\qquad 0,   \qquad  \ \ \ \  \mathrm{otherwise.}
\end{array}\right.
\end{equation}
where $GT$ denotes the ground-truth box.

\subsection{Adaptive Points Assessment and Assignment}
Due to the lack of direct supervision, learning high-quality points is essential to capture the geometric features adaptively for densely-packed and arbitrarily-oriented objects in aerial images. To this end, we propose an effective assessment and assignment scheme to measure the quality of learned points, which towards assigning the representative samples of adaptive points as the positive samples at the training stage. 

\textbf{Quality Measure of Adaptive Points.}
Firstly, we define a quality measure $Q$ to assess the learned adaptive points from four aspects, including classification and localization ability ${Q_{cls}}$, ${Q_{loc}}$, orientation alignment ${Q_{ori}}$, and point-wise correlation ${Q_{poc}}$ for each oriented point set. Thus, $Q$ is derived as below:
\begin{equation}
Q = {Q_{cls}} +  \mu_{1}{Q_{loc}} + \mu_{2} {Q_{ori}}  + \mu_{3} {Q_{poc}}
\label{quality}
\end{equation}
The classification ability $Q_{cls}$ of a point set $\mathcal{R}_i$ directly reflects its classification confidence $\mathcal{R}_i^{cls}(\theta )$, where the corresponding classification loss $\mathcal{L}_{cls}$ measures the compatibility of the points feature with ground-truth class label $b_j^{cls}$. We define $Q_{cls}$ as follows:
\begin{equation}
{Q_{cls}}({\mathcal{R}_i}, {b_j}) = {\mathcal{L}_{cls}}(\mathcal{R}_i^{cls}(\theta ),b_j^{cls})
\end{equation}

To evaluate the compatibility of the points position with the ground-truth $b_j^{loc}$, we employ the localization loss as the quality assessment measure, which is based on the IoU transformation. It indicates the spatial alignment when the center of a point set is near to the object's geometric center. Therefore, ${Q_{loc}}$ is defined as below:


\begin{equation}
{Q_{loc}}({\mathcal{R}_i},{b_j}) = {\mathcal{{L}}_{loc}}(OB_i^{loc}(\theta ),b_j^{loc})
\end{equation}

Since ${Q_{loc}}$ can be regarded as a measure of spatial location distance, it is insensitive to the orientation variations, especially for the square-like objects in the aerial images. To account for the orientation alignment, we employ  Chamfer distance~\cite{ChamferLoss_CVPR2017} to assess the difference in orientation between the predicted point set and ground-truth box contour points. We firstly adopt the $\textit{MinAeaRect}$ conversion function to obtain four spatial corner points $\{ {v_1},{v_2},{v_3},{v_4}\}$ from the learned point set. Then, an ordered point set ${\mathcal{R}^v}$ (40 points by default) is sampled with the equal interval from two adjacent corner points. Similarly, the points ${\mathcal{R}^g}$ are generated for the ground-truth corner points $\{{g_1},{g_2},{g_3},{g_4}\}$. Therefore, $Q_{ori}$ is defined as follows:
\begin{equation}
{Q_{{\mathop{ori}}}}(\mathcal{R}_i,{{b_j}}) = \mathcal{CD}(\mathcal{R}_{{i}}^v(\theta ),\mathcal{R}_{{b_j}}^g)
\label{qualityequation}
\end{equation}	
where $\mathcal{CD}$ denotes Chamfer distance between the above two group of sampling points: 
\begin{equation}
\begin{split}
{\mathcal{CD}}(\mathcal{R}^v,\mathcal{R}^g) & = \frac{1}{{2n}}\sum\limits_{i = 1}^n {\mathop {\min }\limits_j } {\left\| {(x_i^v,y_j^v) - (x_i^g,y_j^g)} \right\|_2} \\ & + \frac{1}{{2n}}\sum\limits_{j = 1}^n {\mathop {\min }\limits_i } {\left\| {(x_i^v,y_j^v) - (x_i^g,y_j^g)} \right\|_2}
\end{split}
\end{equation}
$(x_i^v,y_j^v) \in {\mathcal{R}^v} $ denotes the sampled points of predicted spatial corner points, and $(x_i^g,y_j^g) \in {\mathcal{R}^g}$ denotes the sampled points generated from ground-truth corner points. 

To measure the point-wise association upon a point set for an oriented object, we extract the point-wise features and employ the cosine similarity between the feature vectors as the correlation measure $Q_{poc}$ for the learned adaptive points. Let $e_{i,k}$ denote the $k$-th point-wise feature vector of $i$-th set of adaptive points. $e_{i,k}^ *$ and $e_i^*$ represent the normalized embedding feature vector and their mean from the $i$-th point set:
\begin{equation}
e_{i,k}^ *  = \frac{{{e_{i,k}}}}{{{{\left\| {{e_{i,k}}} \right\|}_2}}}
\end{equation}

\begin{equation}
e_i^* = \frac{1}{{{N_p}}}\sum\limits_{k = 1} {e_{i,k}^ * } 
\end{equation}
where $N_p$ denotes the num of points in a point set. The default setting is 9. Based on the above notations, $Q_{poc}$ of $i$-th point sets can be formulated as the point-wise feature diversity as below:
\begin{equation}
\begin{split}
{Q_{poc}} &= 1 - \frac{1}{{{N_{p}}}}\sum\limits_k {\cos  < e_{i,k}^ * ,e_i^* > }  \\ &=1{\rm{ - }}\frac{1}{{{N_{p}}}}\sum\limits_{k} {\frac{{{e}_{i,k}^* \cdot {e}_i^*}}{{\left\| {{e}_{i,k}^*} \right\| \times \left\| {{e}_i^*} \right\|}}} 
\end{split}
\end{equation}

\begin{figure}[t]
	\centering
	\begin{subfigure}{0.235\textwidth}
		\includegraphics[width=\textwidth]{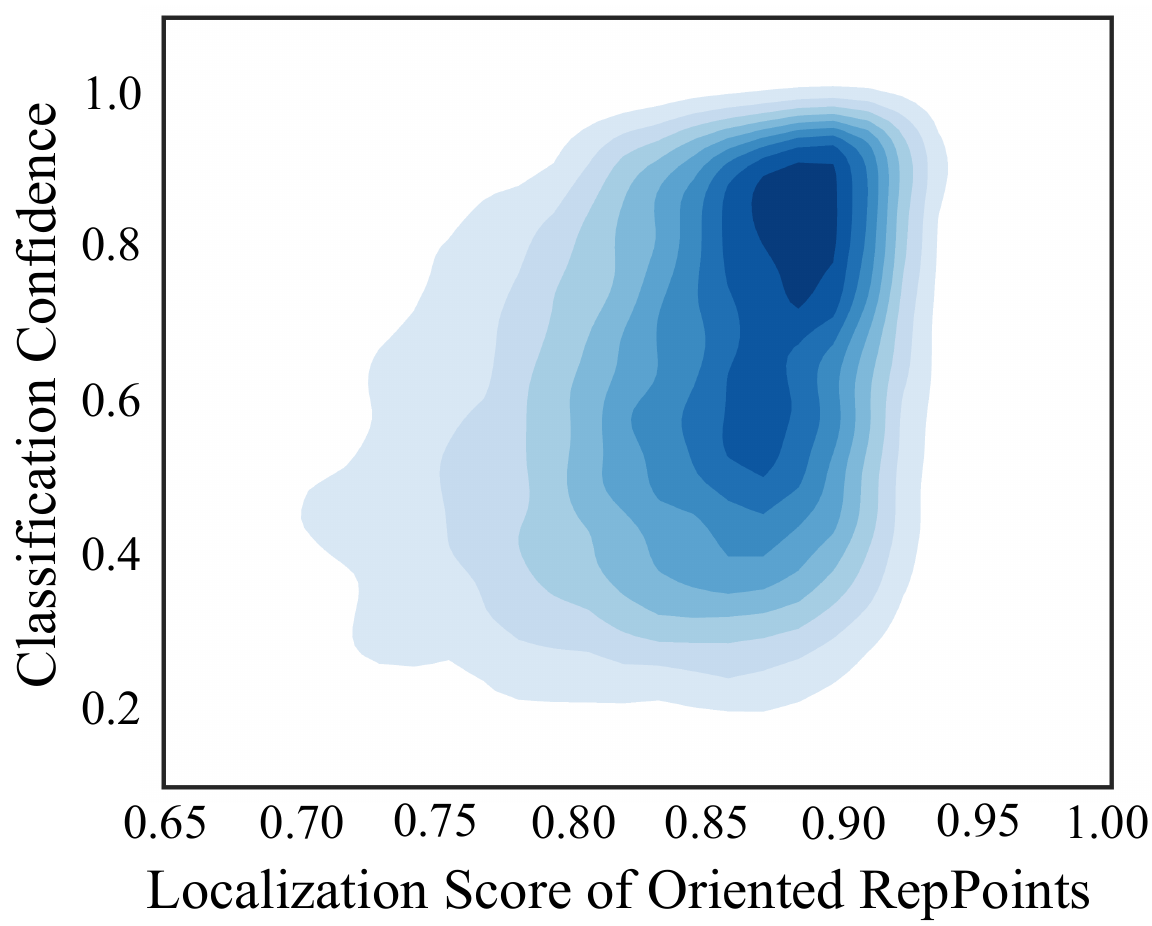}
		\caption{w/o. APAA}
	\end{subfigure}
	\hfill
	\begin{subfigure}{0.235\textwidth}
		\includegraphics[width=\textwidth]{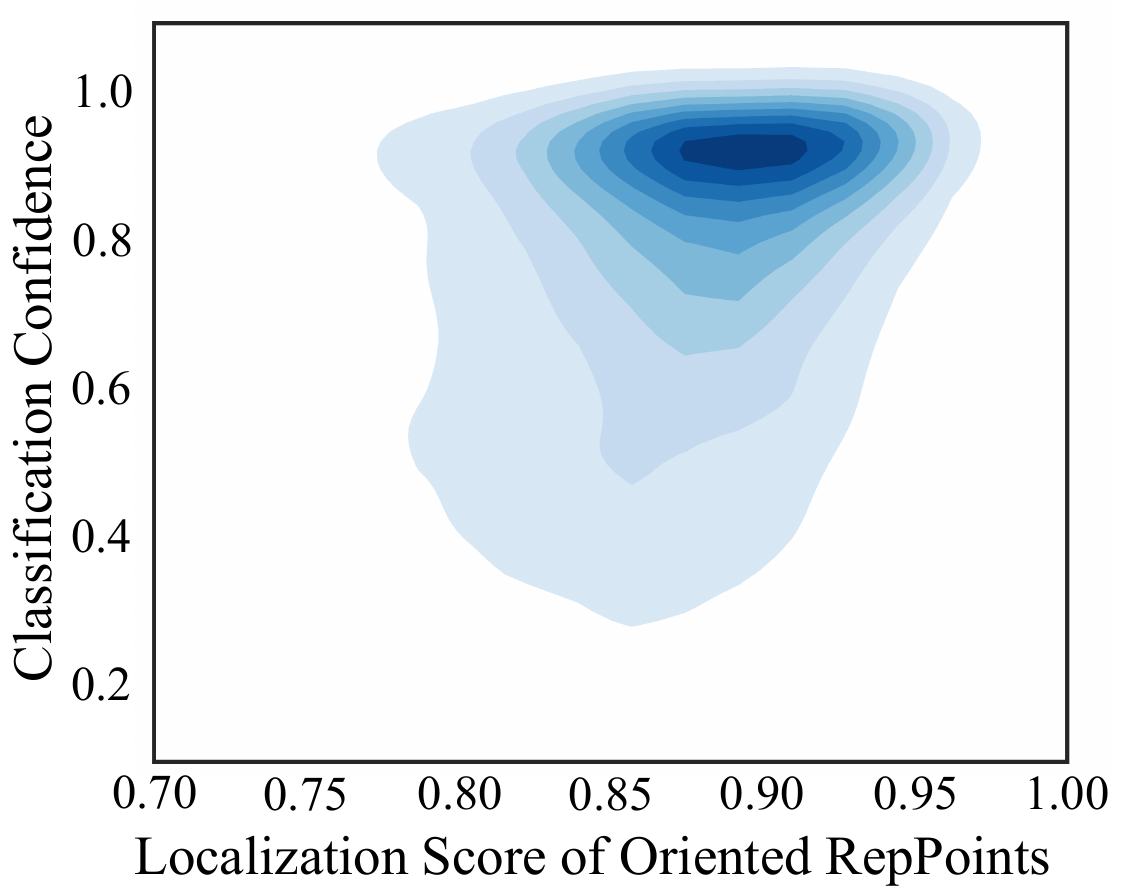}
		\caption{w. APAA}
	\end{subfigure}
	\caption{The correlation between the predicted classification confidence and localization score  of the oriented reppoints with and without APAA scheme.}
	\label{fig:densitymap}
\end{figure}

\textbf{Dynamic $k$ Label Assignment.}
Based on the quality  measure $Q$, we assign the oriented reppoints samples through an efficient and dynamic top $k$ item selection scheme at different iteration. For each object, we sort all the point set samples from the initialization stage according to their quality scores. To retrieve high-quality adaptive point set samples, we set a sampling ratio $\mathbf{\sigma} $ to assign the top $k$ samples at each iteration as the positive samples for training, which is calculated by:
\begin{equation}
k = \sigma {\rm{*}}{N_{t}}
\end{equation}
where $N_{t}$ denotes the total number of point set samples at the initialization stage for each oriented object.

During the training, the points assigner~\cite{DBLP:conf/iccv2019/RepPoint} is used to get the sample assignment of center points at the initialization stage. At the refinement stage, the proposed adaptive points assessment and assignment (APAA) scheme is used to select the high-quality points samples according to the quality measure $Q$. Only the selected positive point sets are assigned with the ground-truth bounding box of target. As shown in Fig.~\ref{fig:densitymap}, APAA scheme enables the detector to predict the high-quality oriented reppoints for improving both classification confidence and localization scores. It is worthy of mentioning that the presented scheme is only used for training, which does not incur the computational load at the inference stage.

\section{Experiments}

\subsection{Evaluation Testbed}
DOTA~\cite{DBLP:conf/cvpr2018/DOTA} is a large-scale dataset to evaluate the detection performance of oriented objects in aerial images, which contains 2806 images, 188,282 instances and 15 categories with a variety of orientations, scales, and shapes. The training set has 1411 images while the validation set contains 458 images. Testing set is made of 937 images. The image sizes range from 800 $\times$ 800 to 4000 $\times$ 4000. In our experiments, both the training set and validation set are employed to train the proposed detector, and the testing set without annotations is used for evaluation. We crop the original images into the patches of 1024 $\times$ 1024 with a stride of 824. At the training stage, we randomly resize and flip the images 
to avoid overfitting. 



HRSC2016~\cite{DBLP:conf/icpram/HRSC2016} contains a large number of strip-like oriented objects with diverse appearances collected from several famous harbors for ship recognition. The entire dataset has 1061 images ranging from 300 $\times$ 300 to 1500 $\times$ 900. For a fair comparison, the training set (436 images) and validation set (181 images) are employed for training, and the testing set (444 images) is used for evaluation.

UCAS-AOD~\cite{DBLP:conf/icip/UCAS-AOD} has 1510 images with 510 car images and 1000 airplane images. There are 14,596 instances in total. The entire dataset is randomly divided into 755 images for training, 302 images for validation and 453 images for testing with a ratio of 5:2:3. The size of all images is approximately 1280 $\times$ 659.

DIOR-R~\cite{arxiv2021anchor} gives the oriented bounding box annotations based on the DIOR dataset~\cite{ISPRS2020object} for the oriented detection task. There are 23,463 images with the size of 800 $\times$ 800 and 192,518 instances covering 20 object classes. 

\subsection{Implementation Details}
We implement our proposed approach based on both ResNet-50~\cite{resnet50-cvpr2016} and ResNet-101 backbone with FPN~\cite{DBLP:conf/cvpr2017/FPN}. The FPN consists of $P_3$ to $P_7$ pyramid levels in our work. The stochastic gradient descent (SGD) optimizer is used in training. The initial learning rate is set to 0.008 with the warming up for 500 iterations, and the learning rate is decreased by a factor of 0.1 at each decay step. The momentum is set to 0.9, and weight decay is $10^{-4}$. We train the models with 40 epochs, 40 epochs, 120 epochs and 120 epochs for DOTA, DIOR-R, HRSC2016 and UCAS-AOD, respectively. The scale jitter is employed during the training phase. The hyperparameters of focal loss are set to $\alpha $ = 0.25 and $\gamma $ = 2.0. In Eq.~(\ref{overallloss}), we set the balanced weights for each stage $\lambda_1$ = 0.3 and $\lambda_2$ =1.0, empirically. We set $\mu_{1}$ =1.0, $\mu_{2}$ = 0.3 and $\mu_{3}$ = 0.1 for quality evaluation $Q$ in Eq.~(\ref{quality}). A couple of experiments  are performed to choose the appropriate values of the sampling ratio  $\sigma $ in Table~\ref{tab:alpha_analysis}. 

We conduct the experiments on a server with 4 RTX 2080Ti GPUs using a total batch size of 8 (2 images per GPU) for training while a single RTX 2080Ti GPU is employed for inference.

\subsection{Ablation Study}

To examine the effectiveness of each component in our proposed framework, a series of ablation experiments are performed on DOTA dataset with ResNet-50-FPN.

\textbf{Evaluation on oriented conversion functions.}
The conventional point set-based object detector RepPoints~\cite{DBLP:conf/iccv2019/RepPoint} obtains the upright bounding boxes by square conversion function like \textit{min-max}, which cannot deal with the aerial objects having arbitrary orientations. To build a reasonable baseline, we compare the different conversion functions that map the adaptive points into the oriented box during the training and post-processing. Table~\ref{tab:baseline} shows the experimental results. Based on the original RepPoints with \textit{min-max} function both in training and post-processing, it is able to achieve 49.69\% mAP. With the proposed oriented \textit{MinAeraRect} function for post-processing to get the rotated rectangular boxes, it achieves the 53.21\% mAP. With the differentiable oriented \textit{NearestGTCorner} and \textit{ConvexHull} functions, our Oriented RepPoints obtains the 66.97\% mAP and 68.89\% mAP, which shows that the oriented conversion functions are essential to aerial object detection. 

\begin{table}[htbp]
	\centering
	\small
	\setlength{\tabcolsep}{1.0mm}{
		\begin{tabular}{cccc}
			\toprule	
			Methods & training & post-processing & mAP\\
			\midrule
			\multirow{2}*{RepPoints~\cite{DBLP:conf/iccv2019/RepPoint}}
			& \textit{min-max} & \textit{min-max} & 49.69 \\
			~ & \textit{min-max} & \textit{MinAeraRect} & 53.21 \\
			\hline
			\multirow{2}*{Oriented RepPoints}   &\textit{NearestGTCorner} &  \textit{MinAeraRect} & 66.97  \\
			~ &\textit{ConvexHull}  & \textit{MinAeraRect} & \textbf{68.89}  \\
			\bottomrule
	\end{tabular}}
	\caption{{{Comparison with  different conversion functions.}}}
	\label{tab:baseline}
\end{table}

\begin{table}[htbp]
	\centering
	\small
	\setlength{\tabcolsep}{3.0mm}{
		\begin{tabular}{ccc}
			\toprule
			Methods   &  Backbone & mAP \\
			\midrule
			angle-based detector  & ResNet-50-FPN  & 67.50\\
			Oriented RepPoints   & ResNet-50-FPN & \textbf{68.89 (+1.39)}  \\
			angle-based detector  & ResNet-101-FPN & 68.73 \\
			Oriented RepPoints   & ResNet-101-FPN  & \textbf{70.19 (+1.46)}\\
			\bottomrule
	\end{tabular}}
	\caption{{{Comparisons between direct angle-based orientation regression and the Oriented RepPoints for oriented object detection.}}}
	\label{tab:pointsset_and_angle}
\end{table}

\textbf{Comparison with angle-based detectors.}
To examine the effectiveness of adaptive points representation, we compare our approach with the angle-based orientation regression on the anchor-based detector. As in S$^2$A-Net~\cite{s2anet-2020}, the angle-base detector presets one squared anchor for each feature map location at the initialization stage, where the predicted angle-based boxes are regarded as the refined anchors for the
next stage to get the oriented bounding box. Table~\ref{tab:pointsset_and_angle} shows
the results of two detectors with different backbones. The Oriented RepPoints outperforms the angle-based orientation regression with +1.39\% and +1.46\% mAP improvement using ResNet-50-FPN and ResNet-101-FPN backbone, respectively.

\begin{table}[tbp]
	\centering
	\small
	\setlength{\tabcolsep}{1.7mm}{
		\begin{tabular}{cccccc}
			\toprule
			Spatial Constraint  & BD & BR &RA & HC & mAP  \\
			\midrule
			   & 76.85 & 41.72  &67.09 & 41.55 &68.89 \\
			$\checkmark$  &  \textbf{79.99 }  & \textbf{45.33}  &\textbf{71.39} & \textbf{51.87} & \textbf{70.11} \\
			 improvement & +3.14 & +3.61 & +4.30 &  \textbf{+10.02} & +1.22 \\
			\bottomrule
	\end{tabular}}
	\caption{{{Performance evaluation on spatial constraints. BD, BR, RA and HC denote the categories of Baseball Diamond, Bridge, Roundabout and Helicopter, respectively.  }}}
	\label{tab:spatial constraint}
\end{table}  
\begin{table*}[t]
	\centering
	 \scriptsize
	\setlength{\tabcolsep}{1.2mm}{
		\begin{tabular}{ccccccccccccccccc|c}
			\toprule 
			Methods & Backbone & PL& BD& BR& GTF& SV& LV& SH& TC& BC& ST& SBF& RA& HA& SP& HC & mAP \\
			\midrule
			\textit{Single-stage Methods}\\
			RetinaNet-O~\cite{DBLP:conf/iccv2017/FocalLoss} & R-50-FPN &88.67 &77.62 &41.81 &58.17 &74.58 &71.64 &79.11 &90.29 &82.18 &74.32 &54.75 & 60.60 &62.57 &69.67 &60.64 &68.43 \\
			DAL~\cite{aaai2021dynamic} & R-101-FPN & 88.61 & 79.69 & 46.27 & 70.37& 65.89 &76.10 &78.53 &90.84 &79.98 &78.41 &58.71 &62.02 &69.23 &71.32 &60.65 &71.78 \\
			RSDet~\cite{AAAI2020-RSdet} &R-152-FPN & \color{blue}{\textbf{90.10}} &82.00& 53.80 &68.50& 70.20& 78.70& 73.60& 91.20& 87.10& 84.70& 64.30& 68.20& 66.10& 69.30 &63.70& 74.10\\
			R$^3$Det~\cite{AAAI2020r3det} & R-152-FPN & 89.49 & 81.17 &50.53 &66.10 &70.92 &78.66 &78.21 &90.81 &85.26 &84.23 &61.81 &63.77 &68.16 &69.83 &\color{red}{\textbf{67.17}} &73.74\\
			S$^2$A-Net ~\cite{s2anet-2020} & R-50-FPN & 89.11 &82.84 &48.37 &71.11 &78.11 &78.39 &87.25 &90.83 &84.90 &85.64 &60.36 &62.60 &65.26 &69.13 &57.94 &74.12 \\
			R$^3$Det-DCL~\cite{DCLCVPR2021} &R-152-FPN &89.78 &83.95& 52.63 &69.70 &76.84 &81.26 &87.30 &90.81 &84.67 &85.27 &63.50 &64.16 &68.96 &68.79& 65.45 &75.54 \\
			\hline
			
			\textit{Two-stage Methods} \\
			Faster RCNN-O~\cite{DBLP:conf/nips2015/FasterRCNN} & R-50-FPN & 88.44 &73.06 &44.86& 59.09& 73.25& 71.49& 77.11& 90.84& 78.94& 83.90& 48.59& 62.95& 62.18& 64.91& 56.18& 69.05\\
			

			CAD-Net~\cite{TGRS2019-CAD} & R-101-FPN & 87.80 & 82.40 & 49.40 & 73.50 & 71.10 & 63.50 & 76.60 & 90.90 & 79.20 & 73.30 & 48.40 & 60.90 & 62.00 & 67.00 & 62.20 & 69.90 \\  
			SCRDet~\cite{DBLP:conf/iccv2019/SCRDet}& R-101-FPN & 89.98	&80.65	&52.09	&68.36	&68.36	&60.32	&72.41	& 90.85 &\color{red}{\textbf{87.94}}	&\color{red}{\textbf{86.86}}	&\color{blue}{\textbf{65.02}}	&66.68	&66.25	&68.24	&65.21	&72.61 \\
			FAOD~\cite{icip2019FAOD}& R-101-FPN & \color{red}{\textbf{90.21}} &79.58 &45.49 &76.41 &73.18 &68.27& 79.56 &90.83 &83.40 &84.68& 53.40 &65.42 &74.17 &69.69 &64.86& 73.28 \\
			RoI-Trans.~\cite{DBLP:conf/cvpr2019/RoI-Transformer} & R-101-FPN &88.65& 82.60 &52.53& 70.87 &77.93& 76.67& 86.87 &90.71 &83.83& 82.51 &53.95 &67.61 &74.67& 68.75 &61.03 &74.61\\
			
			Gliding Vertex~\cite{TPMI2020-Gliding} & R-101-FPN & 89.64 &\color{blue}{\textbf{85.00}} &52.26 & \color{red}{\textbf{77.34}} &73.01 &73.14 & 86.82 &90.74 &79.02 & \color{blue}{\textbf{86.81}} & 59.55 &\color{red}{\textbf{70.91}} &72.94 & 70.86 & 57.32 & 75.02 \\ 
			MaskOBB~\cite{Remotesensing-maskOBB} & R-50-FPN& 89.61 & \color{red}{\textbf{85.09}}& 51.85& 72.90 &75.28 &73.23 &85.57 &90.37 &82.08& 85.05 &55.73& 68.39& 71.61 &69.87 &\color{blue}{\textbf{66.33}} &74.86\\
			CenterMap~\cite{TGRS2021-Centermap} & R-50-FPN &88.88 &81.24 &53.15 &60.65 &78.62 &66.55 &78.10 &88.83& 77.80 &83.61 &49.36& 66.19 &72.10 &72.36 &58.70 &71.74\\
			ReDet~\cite{cvpr2021redet} & ReR-50-ReFPN~\cite{cvpr2021redet} &88.79  &82.64 &53.97 &74.00 &78.13 &\color{red}{\textbf{84.06}} &\color{red}{\textbf{88.04}} &90.89 &\color{blue}{\textbf{87.78}} &85.75 &61.76& 60.39 &\color{red}{\textbf{75.96}} &68.07 &63.59 &76.25\\
			Oriented R-CNN~\cite{ICCV2021_orientedrcnn} & R-101-FPN &88.86 &83.48 &55.27 &\color{blue}{\textbf{76.92}} &74.27 &82.10 &87.52 &\color{blue}{\textbf{90.90}} &85.56 &85.33 &\color{red}{\textbf{65.51}} &66.82 &74.36 &70.15 &57.28 &76.28\\
			\hline
			\textit{Anchor-free Methods}\\
			CenterNet-O~\cite{centernet2019objects} & DLA-34~\cite{centernet2019objects} & 81.00 &64.00 &22.60& 56.60 &38.60 &64.00 &64.90 &90.80& 78.00 &72.50 &44.00 &41.10 &55.50& 55.00 &57.40 &59.10 \\
			PIoU~\cite{eccv2020piou} & DLA-34 & 80.90 &69.70 &24.10 &60.20 &38.30 &64.40 &64.80 &90.90 &77.20 &70.40 &46.50 &37.10 &57.10 &61.90& 64.00 & 60.50 \\
			O$^2$-DNet~\cite{O2-DNET-ISPRS2020} & H-104~\cite{h104-cvprw2017} &89.31 & 82.14 &47.33 & 61.21 & 71.32 & 74.03 & 78.62 & 90.76 & 82.23 & 81.36 & 60.93 & 60.17 & 58.21 & 66.98 & 61.03 & 71.04 \\
			DRN~\cite{CVPR2020-DRN} & H-104& 89.71 & 82.34 & 47.22 & 64.10 & 76.22 & 74.43 & 85.84 & 90.57 & 86.18 & 84.89 & 57.65 & 61.93 & 69.30 & 69.63 & 58.48 & 73.23 \\ 
			CFA~\cite{CVPR2021beyond}  & R-101-FPN  &  89.26 & 81.72 &51.81 &67.17 &79.99  &78.25& 84.46 &90.77 &83.40 &85.54 &54.86 &67.75 &73.04 &70.24& 64.96& 75.05 \\
			Oriented RepPoints & R-50-FPN & 87.02 &  83.17 & 54.13 & 71.16 & \color{blue}{\textbf{80.18}} & 78.40 & 87.28 & \color{red}{\textbf{90.90}} & 85.97 & 86.25 & 59.90 & \color{blue}{\textbf{70.49}} & 73.53 & 72.27 & 58.97 & 75.97 \\
			
			Oriented RepPoints & R-101-FPN &89.53& 84.07 &\color{red}{\textbf{59.86}} & 71.76 & 79.95 & 80.03 & 87.33 & 90.84 &87.54 &85.23 & 59.15 &66.37 &75.23 & \color{red}{\textbf{73.75}} &57.23 & \color{blue}{\textbf{76.52}} \\	
			Oriented RepPoints& Swin-T-FPN  &  89.11 & 82.32 & \color{blue}{\textbf{56.71}} & 74.95 & \color{red}{\textbf{80.70}} &\color{blue}{\textbf{83.73}} & \color{blue}{\textbf{87.67}} & 90.81 & 87.11 & 85.85 & 63.60 & 68.60 & \color{blue}{\textbf{75.95}} & \color{blue}{\textbf{73.54}} & 63.76 & \color{red}{\textbf{77.63}} \\
			\bottomrule
	\end{tabular}}
	\caption{Comparison with state-of-the-art methods on DOTA dataset. All reported results are performed on the single-scale DOTA dataset. The results with \textcolor{red}{red} color denote the best results and with \textcolor{blue}{blue} color present the second-best results in each column. '-O' means the detection results with oriented bounding box (the same below).}
	\label{tab:DOTAResults}
\end{table*}

\textbf{Evaluation on spatial constraint.}
To investigate the effectiveness of spatial constraint, we compare it against the baseline method without using it. Table~\ref{tab:spatial constraint} shows the experimental results. It can be observed that our proposed spatial constraint is very effective, especially for the aerial objects with weak feature representation such as HC (Helicopter), and the objects similar to the background, e.g. BD (Baseball Diamond), BR (Bridge) and RA (Roundabout). This is because the spatial constraint enforces the adaptive points on their owner instance object.

\begin{table}[htbp]
	\centering
	\small
	\setlength{\tabcolsep}{2.8mm}{
		\begin{tabular}{cccccc}
			\toprule
			 \multicolumn{6}{c}{Quality Measure $Q$ for Adaptive Points.}  \\ 
			\midrule
			$Q_{cls}$  &  &  $\checkmark$ & $\checkmark$ & $\checkmark$   & $\checkmark$   \\
			$Q_{loc}$ &  &  & $\checkmark$ & $\checkmark$ & $\checkmark$ \\
			$Q_{ori}$  & &  & & $\checkmark$ & $\checkmark$ \\ 
		    $Q_{poc}$ & &  & &  & $\checkmark$  \\ 
			mAP & 70.11 & 72.34 & 74.46  & 75.32 & \textbf{75.97}   \\
			\bottomrule
	\end{tabular}}
	\caption{{{Performance evaluation on different settings of quality measure $Q$ in APAA scheme.}}}
	\label{tab:settingofselection}
\end{table}

\begin{table}[htbp]
	\centering
	\small
	\setlength{\tabcolsep}{4mm}{
		\begin{tabular}{ccccc}
			\toprule
			$\mathbf{\sigma}$  & 0.2  & 0.3 & 0.4 & 0.5\\
			\midrule
			mAP & 75.45 & 75.34 &  \textbf{75.97}  & 75.67 \\
			\bottomrule
	\end{tabular}}
	\caption{{{Evaluation on various $\mathbf{\sigma}$ in the dynamic top $k$ assignment of APAA scheme.}}}
	\label{tab:alpha_analysis}
\end{table}
\begin{table}[htbp]
	\centering
	\small
	\setlength{\tabcolsep}{0.85mm}{
		\begin{tabular}{cccccc}
			\toprule
			Methods  & Max-IoU~\cite{DBLP:conf/nips2015/FasterRCNN} & ATSS~\cite{cvpr2020atss} &  	PAA~\cite{eccv2020paa}  & CFA~\cite{CVPR2021beyond} & 
			APAA\\
			\midrule
		    mAP  & 70.11 & 72.87  & 74.62 & 74.89  & \textbf{75.97}  \\
			\bottomrule
	\end{tabular}}
	\caption{{{Comparisons with different samples assignment methods on Oriented RepPoints detector.}}}
	\label{tab:labelassignment}
\end{table}



\textbf{APAA scheme for adaptive points learning.}
To study the proposed APAA scheme for adaptive points learning, we first report the performance of the quality measure term-by-term. Table~\ref{tab:settingofselection} gives the results of different settings on quality assessment measure $Q$. The detection result is progressively improved, and the proposed approach achieves the best performance with 75.97\% mAP and +5.86\% gains using all four terms.  
It indicates that the quality assessment measure is effective to reflect the quality of adaptive points for aerial object detection. In APAA scheme, the number of assigned adaptive points samples is determined by the sampling ratio $\sigma$. 
As shown in Table~\ref{tab:alpha_analysis}, the model achieves the best performance when $\sigma = 0.4$. Additionally, we compare our APAA scheme with other sample assignment schemes for training the proposed detector, including Max-IoU~\cite{DBLP:conf/nips2015/FasterRCNN}, ATSS~\cite{cvpr2020atss}, PAA~\cite{eccv2020paa} and CFA~\cite{CVPR2021beyond}. As illustrated in Table~\ref{tab:labelassignment}, our APAA scheme achieves the best performance without the complicated operations, which demonstrates that our proposed APAA is effective for adaptive points learning.
\begin{figure*}[t]
	\centering
	\includegraphics[width=0.92\linewidth]{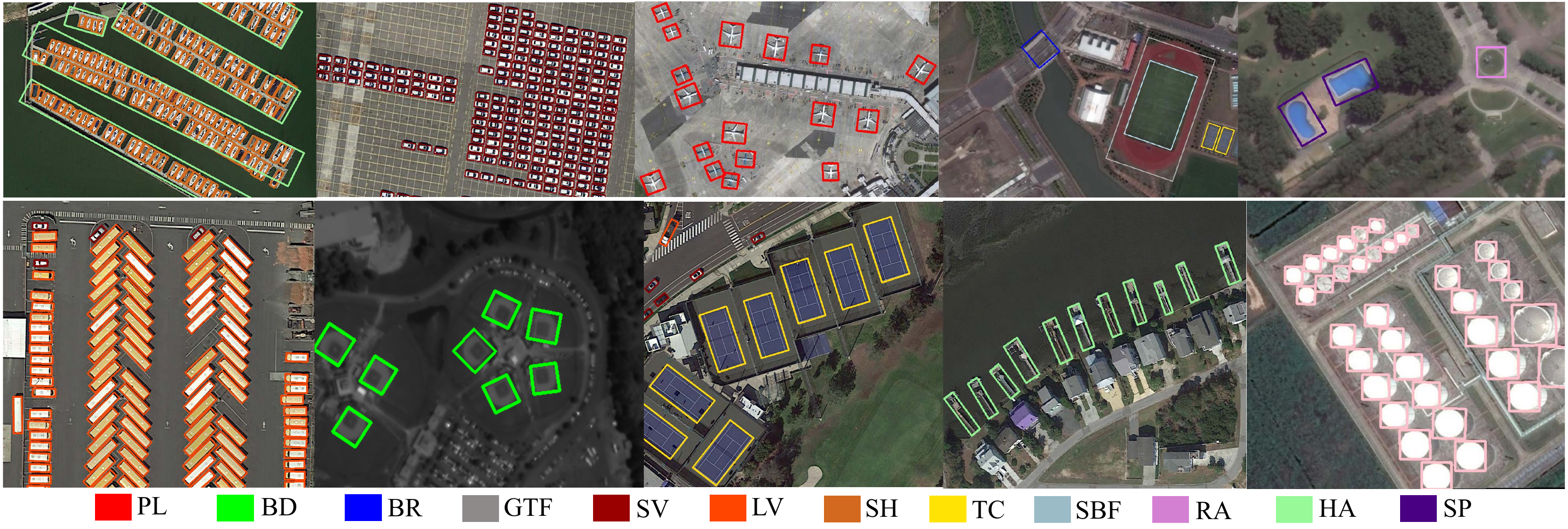}
	\caption{Example detection results of Oriented RepPoints on DOTA test set.}
	\label{dota_visule_results}
\end{figure*}
\subsection{Comparison with the State-of-the-art methods}
\textbf{Results on DOTA.} We report full experimental results of single scale to make a fair comparison with the previous methods on DOTA dataset. With ResNet-50-FPN and ResNet-101-FPN as the backbones, our method obtains 75.97\% and 76.52\% mAP, respectively. It outperforms other methods with the corresponding backbones. Using the tiny version of Swin-Transformer~\cite{swiniccv2021} (Swin-T-FPN) as backbone with the random rotation and HSV transformation,  we achieve the best performance with 77.63\% mAP. Fig.~\ref{dota_visule_results} shows some visual results on DOTA test set.

\textbf{Results on HRSC2016.} To make a comprehensive comparison on HRSC2016, we report the results with both VOC2007 and VOC2012 metrics. Table~\ref{tab:HRSC2016Results} shows the experimental results.  Our Oriented RepPoints achieves the best performance under VOC2012 metric and the second-best under VOC2007 metric with ResNet-50-FPN backbone. 

\textbf{Results on UCAS-AOD.} UCAS-AOD datasets contains a large number of small objects with complex surrounding scenes.  Table~\ref{tab:UCASAODResults} shows the evaluation results with the recent methods on UCAS-AOD dataset. Our presented method achieves the best performance of 90.11\% mAP.

\textbf{Results on DIOR-R.} DIOR-R datasets consists of 20 classes of aerial objects. Compared with the recent methods on this dataset, we achieve the best performance with 66.71\% mAP and outperform other methods, as shown in Table~\ref{tab:DIOR-R-results}.

\begin{table}[tbp]
	\centering
	\footnotesize
	\setlength{\tabcolsep}{2.0mm}{
		\begin{tabular}{cccc}
			\toprule
			Methods &Backbone & mAP$_{50}(07)$ & mAP$_{50}(12)$\\
			\midrule
			R$^2$CNN~\cite{arxiv2017r2cnn}  & R-101-FPN  & 73.07 & 79.73 \\
			RRD~\cite{CVPR2018-RRD}  & VGG16 & 84.30 & - \\ 
			RoI-Trans.~\cite{DBLP:conf/cvpr2019/RoI-Transformer} & R-101-FPN   &86.20 &- 	\\
			CenterNet-O~\cite{centernet2019objects} & DLA-34 & 87.89 & - \\
			Gliding Vertex~\cite{TPMI2020-Gliding} &R-101-FPN   &88.20 &-  \\
			DRN~\cite{CVPR2020-DRN} & H-104  & - & 92.70 \\ 
			CenterMap-Net~\cite{TGRS2021-Centermap}  & R-50-FPN  & - &92.80  \\
			RetinaNet-O~\cite{DBLP:conf/iccv2017/FocalLoss} & R-101-FPN & 89.18 & 95.21 \\ 
			PIOU~\cite{eccv2020piou} & DLA-34 & 89.20 & - \\
			MFIAR-Net~\cite{DBLP:journals/sensors/MFIAR-Net}  & R-101-FPN & 89.81 & -\\
			R$^3$Det~\cite{AAAI2020r3det}  & R-101-FPN &89.26 & 96.01 \\
			R$^3$Det-DCL~\cite{DCLCVPR2021} & R-101-FPN & 89.46 & 96.41 \\
			FPN-CSL~\cite{CSL_ECCV2020}  & R-101-FPN & 89.62 & 96.10 \\
			DAL~\cite{aaai2021dynamic} & R-101-FPN & 89.77 & - \\
			S$^2$A-Net~\cite{s2anet-2020} & R-101-FPN& 90.17 & 95.01   \\ 
			Oriented R-CNN~\cite{ICCV2021_orientedrcnn}  & R-50-FPN & \textbf{90.40} & 96.50\\
			Oriented RepPoints  & R-50-FPN & 90.38 & \textbf{97.26} \\ 
			\bottomrule
	\end{tabular}}
	\caption{{{Results on HRSC2016 test set. mAP$(07)$ and mAP$(12)$ represent the results under VOC2007 and VOC2012 mAP metrics.}}}
	\label{tab:HRSC2016Results}
\end{table}

\begin{table}[tbp]
	\centering
	\footnotesize
	\setlength{\tabcolsep}{4.0mm}{
		\begin{tabular}{cccc}
			\toprule
			Methods & Car &  Airplane & mAP \\
			\midrule
			YOLOv3-O~\cite{ARXIV-yolov3} & 74.63 & 89.52  & 82.08  \\
			RetinaNet-O~\cite{DBLP:conf/iccv2017/FocalLoss} & 84.64 & 90.51 & 87.57  \\
			Faster R-CNN-O~\cite{DBLP:conf/cvpr2018/DOTA} &86.87 & 89.86 & 88.36   \\
			RoI Trans.~\cite{DBLP:conf/cvpr2019/RoI-Transformer} & 87.99 & 89.90 & 88.95  \\ 
			DAL~\cite{aaai2021dynamic} & 89.25 & 90.49 & 89.87 \\
			Oriented RepPoints & \textbf{89.51}  & \textbf{90.70}  &  \textbf{90.11} \\
			\bottomrule
	\end{tabular}}
	\caption{{{Performance comparisons on UCAS-AOD dataset.}}}
	\label{tab:UCASAODResults}
\end{table}
\begin{table}[htbp]
    \centering
    \footnotesize
    \setlength{\tabcolsep}{0.95mm}{
        \begin{tabular}{cccc}
        \toprule
         Methods &   RetinaNet-O~\cite{DBLP:conf/iccv2017/FocalLoss} & Faster RCNN-O~\cite{DBLP:conf/nips2015/FasterRCNN} & Gliding Vertex~\cite{TPMI2020-Gliding} \\
         mAP &  57.55 & 59.54 & 60.06 \\
         \midrule
         Methods& RoI-Trans.~\cite{DBLP:conf/cvpr2019/RoI-Transformer} & AOPG~\cite{arxiv2021anchor} & Oriented RepPoints \\
         mAP & 63.87 & 64.41 & \textbf{66.71} \\
         \bottomrule
        \end{tabular}}
	    \caption{Detection accuracy on DIOR-R dataset. All experimental results are performed with ResNet-50-FPN backbone.}
	    \label{tab:DIOR-R-results}
\end{table}
\begin{table}[htbp]
    \centering
    \scriptsize
    \setlength{\tabcolsep}{0.4mm}{
        \begin{tabular}{ccccc}
        \toprule
         Methods &   RetinaNet-O~\cite{DBLP:conf/iccv2017/FocalLoss} & Faster RCNN-O~\cite{DBLP:conf/nips2015/FasterRCNN} & RoI-Trans.~\cite{DBLP:conf/cvpr2019/RoI-Transformer} &
         S$^2$A-Net~\cite{s2anet-2020}  \\
         mAOE &  9.53 & 6.01 & 6.35 & 10.43 \\
         \midrule
         Methods&  ReDet~\cite{cvpr2021redet}
         & Oriented R-CNN~\cite{ICCV2021_orientedrcnn} & Oriented RepPoints \\
         mAOE & 6.35 & 7.53 & \textbf{5.93} \\
         \bottomrule
        \end{tabular}}
	    \caption{Comparison on orientation errors with DOTA. All experiments are performed with \textit{train} set for training, \textit{val} set for testing.}
	    \label{tab:mAOE-results}
\end{table}

\subsection{Evaluation on Orientation Accuracy}
We further conduct experiment to evaluate the orientation accuracy of an oriented detector on DOTA dataset with ResNet-50-FPN backbone. We employ the mean Average Orientation Error (mAOE$^\circ$) on all categories as the evaluation metric. As shown in Table~\ref{tab:mAOE-results}, our proposed approach obtains the smallest orientation errors, which demonstrates that our point set-based approach is effective for precise oriented object detection, comparing to the conventional orientation regression-based methods.

\section{Conclusion}
This paper proposed an effective aerial object detector by taking advantage of the adaptive points as a fine-grained representation, which is able to capture the key geometric features for arbitrary-oriented, cluttered and non-axis aligned targets. To effectively learn the adaptive points, we introduced the quality assessment and sample assignment scheme to measure and select the high-quality points samples for training. Furthermore, a spatial constraint is used to penalize the points outside the oriented box for robust adaptive points learning. The extensive experiments have been performed on four testbeds, whose promising results demonstrate the efficacy of our proposed approach.


%

\section*{Acknowledgments}
This work is supported by National Natural Science Foundation of China under Grants (61831015).








{\small
\bibliographystyle{ieee_fullname}
\bibliography{egbib}
}

\end{document}